\documentclass[journal]{article}
\usepackage{cite}
\usepackage{algorithm}
\usepackage{algorithmic}
\usepackage{amsmath}
\usepackage{amssymb}
\usepackage{multirow} 
\usepackage{subfig}
\usepackage{amsthm}
\usepackage{graphicx}

\usepackage{url}
\renewcommand{\vec}[1]{\mathbf{#1}}
\usepackage{array,tabularx}
\usepackage{enumitem}
\newcolumntype{C}[1]{>{\centering\arraybackslash}m{#1}}
\usepackage{xcolor}

\begin{document}

\title{Robust Graph Learning from Noisy Data}

\author{Zhao~Kang\thanks{Z. Kang, H. Pan, and Z. Xu are with the School of Computer Science and Engineering, University of Electronic Science and Technology of China, Chengdu, Sichuan 611731, China. E-mail: \{zkang,~zlxu\}@uestc.edu.cn.}, Haiqi~Pan,
       Steven C.H. Hoi\thanks{S. C.H. Hoi is with School of Information Systems, Singapore Management University,Singapore, Singapore Singapore 17890.
E-mail: {chhoi@smu.edu.sg}},
        ~Zenglin~Xu}
\maketitle

\begin{abstract}
Learning graphs from data automatically has shown encouraging performance on clustering and semisupervised learning tasks. However, real data are often corrupted, which may cause the learned graph to be inexact or unreliable. In this paper, we propose a novel robust graph learning scheme to learn reliable graphs from real-world noisy data by adaptively removing noise and errors in the raw data. We show that our proposed model can also be viewed as a robust version of manifold regularized robust PCA, where the quality of the graph plays a critical role. The proposed model is able to boost the performance of data clustering, semisupervised classification, and data recovery significantly, primarily due to two key factors: 1) enhanced low-rank recovery by exploiting the graph smoothness assumption, 2) improved graph construction by exploiting clean data recovered by robust PCA. Thus, it boosts the clustering, semi-supervised classification, and data recovery performance overall. Extensive experiments on image/document clustering, object recognition, image shadow removal, and video background subtraction reveal that our model outperforms the previous state-of-the-art methods.
\end{abstract}


%

\section{Introduction}
Graph structure is widely used to represent data in real-world. Consequently, graph construction becomes a fundamental problem in machine learning, pattern recognition, and data mining \cite{peng2017nonnegative,kang2018unified,huang2018auto,kang2019similarity,peng2017integrating}. Moreover, it is crucial to the performance of many algorithms. In this paper, we focus on two basic applications: clustering \cite{chen2012fgkm,yang2017discrete,wang2016learning,chang2015convex,huang2018self,Ren2017WOEC} and semisupervised classification \cite{wang2008semi,li2016graph,Berton2015,nie2016parameter,chao2019semi,li2018collaborative}. Due to their promising performance, a variety of graph-based clustering and classification methods have been developed during the past decades. 

Clustering partitions the data objects into different groups based on certain property \cite{peng2017subspace,Ren2018SDEC,chen2017dnc}. As a classical task, a number of clustering techniques have been developed. Among them, two most successful techniques are $k$-means and spectral clustering. Compared with $k$-means and its extensions \cite{huang2018robust,du2015robust,LiuWYZZ13,chen2013twkm,liu2017spectral
,peng2018integrate,liu2019pami}, spectral clustering method presents more capability in detecting complex structures of data \cite{von2007tutorial,ng2002spectral,wang2018multiview,kang2017kernel,yang2015multitask}. It works by embedding the data point into a vector space that is spanned by the spectrum of the graph matrix. Therefore, the quality of the graph is crucial to the performance of spectral clustering algorithm. Typically, the graph is constructed by some computation criterion from raw data \cite{yu2015learning}, where the Gaussian kernel function or $k$-nearest-neighbor approaches are often adopted. However, how to choose an appropriate scaling factor $\sigma$ or a proper neighbor number $k$ are still open questions. More importantly, both of them highly influence the clustering results \cite{zelnik2004self}. 

On the other hand, semisupervised learning deals with the situation that a small portion of samples has been labeled \cite{zhu2018yotube}. In specific, semisupervised classification aims to predict labels for a large number of unlabeled data points based on the relationships among the data points. A lot of work on this task is based on graph regularizer. In general, the first step is to create a graph with the labeled and unlabeled samples as the vertices and with the edge weights encoding the similarity between the samples. Then, the missing labels are inferred based on the local smoothness over the graph, i.e., the more similar two objects are, the more likely they have similar labels. Consequently, this approach's performance depends greatly upon the edge weights of the graph. Although label inference has been thoroughly investigated, graph construction has attracted much less attention until recent years \cite{Berton2015}. There is still no typically used strategy to address this graph construction challenge. 

Recently, learning graphs from data automatically has drawn significant attention \cite{zhan2017graph,li2017multi,peng2017constructing,zhu2018dehazegan,li2018self,zhao2015automatic,kang2018low,Peng2017Connections}. By this means, the most informative neighbors for each data point are automatically selected and it is free of similarity measure metrics, which are often data-dependent and sensitive to noise and outliers \cite{huang2015new}. As a result, the resulted accuracy is much higher than that of  traditional manually constructed graphs. Generally speaking, it can be classified into two categories. The first approach is based on self-expressiveness. Basically, it represents every data point by a linear combination of all other data points and the achieved coefficient matrix is seen as the graph matrix. It has demonstrated impressive performance in a number of applications, including clustering \cite{kang2017twin,liu2013robust,elhamifar2013sparse,peng2018structured}, semisupervised learning \cite{zhuang2012non,kang2018self}. 

The other one is adaptive neighbors approach. It builds graph by assigning a probability for each data point as the neighborhood of another data point and the learned probability is treated as the similarity between two data points \cite{nie2014clustering}. This approach is supposed to capture the local structure of data. Due to its simplicity and effectiveness, it has been widely utilized in various domains, such as clustering \cite{nie2014clustering}, nonnegative matrix factorization \cite{dacheng2017}, feature selection \cite{du2015unsupervised,LiuWZYL14}, multiview learning \cite{nie2017multi}. However, all these work ignore the fact that the graph is learned from the raw data, which is often contaminated in reality. Consequently, the achieved graph might be inexact or sub-optimal and fails to characterize the true relationships between data points.

In this work, we propose a novel robust graph learning scheme to learn robust and reliable graphs from real data. Specifically, we decompose the original data into a low-rank matrix $D$ (``clean data") and a sparse matrix $E$ (``noise/errors"). Then we can build the graph $S$ on the clean data $D$ by using adaptive neighbors approach. Instead of following a two steps approach, we propose a joint learning scheme to optimize $S$ and $D$ simultaneously, such that they are mutually enhanced via an alternating optimization approach. Therefore, besides constructing graph for clustering, our model simultaneously recovers the low-rank clean data, i.e., functioning as robust principle component analysis (RPCA) \cite{candes2011robust}. In summary, the main contributions of our work are:
 
\begin{itemize}
\item{We propose a robust graph learning scheme by decomposing the raw data into a clean part and an error part. Rather than adopting a two steps approach, we develop a unified model and jointly learn the graph as well as the clean data. To the best of our knowledge, this is the first work to investigate robust adaptive neighbor graph construction. }
\item{We show that our proposed method can significantly boost a graph Laplacian regularized RPCA model. Unlike many other heuristic methods for constructing graph Laplacian directly from noisy data, our graph Laplacian is adaptively built from clean data. Benefiting from the incorporation of graph smoothness assumption, the quality of the recovered low-rank matrix is enhanced.    }
\item{Extensive experiments on face clustering, document clustering, face/object recognition, face image shadow removal, and background separation demonstrate the effectiveness of the proposed technique.}
\end{itemize}

In the remainder of this paper, we introduce the graph learning in Section \ref{revist}. Then a robust graph learning technique is proposed in Section \ref{robust}, where the details of the algorithm are also provided. Section \ref{clustering} presents experimental evaluation on clustering task. The semisupervised application is discussed in Section \ref{class}. We compare the data recovery effects in Section \ref{recovery}. Finally, conclusions are drawn in Section \ref{conclusion}.

\section{Adaptive Neighbor Graph Learning Revisited}\label{revist}
 We first introduce the notations that are used throughout the rest of the paper. Given a data matrix $X\in\mathcal{R}^{m \times n}$ with $m$ features and $n$ samples, we denote its $(i,j)$-th element and $i$-th column as $x_{ij}$ and $x_i$, respectively. The $\ell_1$-norm of $X$ is denoted by $\|X\|_1=\sum_{ij}|x_{ij}|$. The definition of $X$'s nuclear norm is $\|X\|_*=\sum_i\sigma_i$, where $\sigma_i$ is the $i$-th singular value of $X$. $\vec{1}$ is a vector with all entries as 1. We use $0\leq S\leq 1$ to represent all elements in $S$ are in the range of $[0,1]$. $(s)_+=\textit{max}(s,0)$.

As locality preserving projection (LPP) \cite{he2004locality} does, we can assign a probability $s_{ij}$ for $x_j$ as the neighborhood of $x_i$. Thus, $s_{ij}$ characterizes the similarity between $x_i$ and $x_j$ in some sense. Smaller distance $\|x_i-x_j\|^2$ indicates that $x_i$ and $x_j$ are quite similar, thus bigger value $s_{ij}$ is. To achieve graph $S$, we can solve the following problem:
\begin{equation}
\begin{split}
&\min_{s_i} \sum_{j=1}^n (\frac{1}{2}\| x_i-x_j\|^2s_{ij}+\gamma s_{ij}^2), \\
&\quad\hspace{.2cm} s.t. \hspace{.1cm} s_i^T\vec{1}=1, \hspace{.1cm}  0\leq s_{ij}\leq 1,
\end{split}
\label{slocal}
\end{equation}
where $\gamma$ is a trade-off parameter. By defining graph Laplacian matrix $L=D-\frac{S+S^T}{2}$, where $D$ is a diagonal matrix with $d_{ii}=\sum_j \frac{s_{ij}+s_{ji}}{2}$, problem (\ref{slocal}) becomes:
\begin{equation}
\begin{split}
&\min_{S} Tr(XLX^T)+\gamma \|S\|_F^2, \\
&\quad\hspace{.2cm} s.t. \hspace{.1cm}S\vec{1}=\vec{1}, \hspace{.1cm}  0\leq S\leq 1.
\end{split}
\label{local}
\end{equation}

By optimizing above problem, one can learn $S$ adaptively from the data. Compared with LPP, where neighborhood relationship is predetermined, the structure is automatically learned from the data in Eq. (\ref{local}). This strategy has attracted great attention recently. \cite{nie2014clustering} performed clustering based on the learned graph. \cite{dacheng2017} employed the obtained graph to construct the Laplacian graph used in graph-based Nonnegative Matrix Factorization (NMF). \cite{du2015unsupervised,cai2018image} presented a feature selection method by using the adaptive graph to preserve the local manifold structure. \cite{nie2017multi} extended this graph learning strategy to multiview data. \cite{wang2017fast} proposed to use Eq. (\ref{local}) to construct Anchor graph for large-scale hyperspectral image clustering. There are also many other applications of Eq. (\ref{local}) that are beyond the scope of this paper. 

However, one big issue is never noticed: it is highly sensitive to noisy input data. This is obvious in Eq. (\ref{local}) since $S$ is built from the raw data $X$, which is often corrupted in practice. Consequently, the performance of this approach is deteriorated. To remedy this, we present a principle to robustify graph learning.
\section{Robust Graph Construction}\label{robust}
\subsection{Formulation}
To make the graph learning robust and reliable, we introduce our approach in this section. The core idea is that the observed data matrix is not perfect but corrupted by errors. Thus, instead of operating on original data or performing some data cleaning that precedes the analysis, we assume the data to be decomposed into two parts: the clean data $D$ and the corruptions $E$. Then, the graph learning is performed on clean data. Following the RPCA idea that the clean data is low-rank and corruptions are sparse \cite{candes2011robust,kang2015robust,yang2018identifiability}, we jointly learn the clean data and the graph. Finally, our proposed Robust Graph Construction (RGC) model can be formulated as:
 \begin{equation}
\begin{split}
&\min_{D, E, S} \|D\|_*+\alpha\|E\|_1+\beta Tr(DLD^T)+\gamma \|S\|_F^2, \\
&\quad  s.t.\hspace{.2cm}  X=D+E, \hspace{.1cm}S\vec{1}=\vec{1}, \hspace{.1cm}  0\leq S\leq 1,
\end{split}
\label{rgc}
\end{equation}
where $\alpha$, $\beta$, and $\gamma$ are all trade-off parameters. We will show later that $\gamma$ can be computed based on a reasonable assumption. In the proposed formulation Eq. (\ref{rgc}), graph learning and noise removal are implemented together, such that they can be iteratively boosted. 

If we assume that the graph $S$ is artificially computed from original data $X$, then $L$ is directly available. Finally, $S$ and $L$ are no longer unknowns. Problem (\ref{rgc}) changes to:
  \begin{equation}
\begin{split}
&\min_{D, E} \|D\|_*+\alpha\|E\|_1+\beta Tr(DLD^T), \\
&\quad  s.t.\hspace{.2cm}  X=D+E.
\end{split}
\label{mrpca}
\end{equation}
This is nothing but the RPCA model, except that it incorporates the manifold smoothness regularizer term. It assumes that the low-rank data $D$ itself lies on a smooth manifold. This is called MRPCA model \cite{shahid2015robust}. 

To summarize, compared with existing work in the literature, our proposed objective function in Eq. (\ref{rgc}) enjoys the following advantages:
\begin{itemize}
\item The graph construction is implemented on clean data. Consequently, the graph quality would be improved. This would be helpful for clustering and semisupervised classification.
\item Interestingly, Eq. (\ref{rgc}) can also be treated as an RPCA model. Different from popular RPCA models, it imposes the smooth manifold regularizer. As a result, the low-rank recovery would be enhanced. 
\item Furthermore, not like the existing graph Laplacian construction method, our graph Laplacian is adaptively learned from raw data. Thus, it would be robust to noise and outliers.

\end{itemize}
\subsection{Optimization Algorithm}
To solve Eq. (\ref{rgc}), we first introduce auxiliary variable $Z$ to facilitate the solution of $D$. Then Eq. (\ref{rgc}) can be written in the following equivalent form:
 \begin{equation}
\begin{split}
&\min_{D, E, S, Z} \|D\|_*+\alpha\|E\|_1+\beta Tr(ZLZ^T)+\gamma \|S\|_F^2, \\
&  s.t.\hspace{.2cm}  X=D+E, \hspace{.1cm}S\vec{1}=\vec{1}, \hspace{.1cm}  0\leq S\leq 1,\hspace{.1cm} Z=D.
\end{split}
\label{ergc}
\end{equation}
It can be solved via alternating direction method of multipliers (ADMM). Removing the equality constraints on $X$ and $Z$, we obtain the augmented Lagrangian function as follows:
 \begin{equation}
\begin{split}
&\mathcal{L}(D, E, S, Z, Y_1, Y_2)=\|D\|_*+\alpha\|E\|_1+\beta Tr(ZLZ^T)+\\
&\gamma \|S\|_F^2+\frac{\mu}{2}(\|D+E-X+\frac{Y_1}{\mu}\|_F^2+\|D-Z+\frac{Y_2}{\mu}\|_F^2), \\
&\quad    s.t.\hspace{.2cm}S\vec{1}=\vec{1}, \hspace{.1cm}  0\leq S\leq 1,
\end{split}
\label{lrgc}
\end{equation}
where $\mu>0$ is a penalty parameter and $Y_1$, $Y_2$ are Lagrangian multipliers. We can solve those unknown variables alternatingly, one at each step, while keeping the others fixed.

Step 1: Updating $D$ while fixing other variables. The problem Eq. (\ref{lrgc}) becomes:
\begin{equation}
\min_D \|D\|_*+\mu \|D-H\|_F^2,
\label{solveD}
\end{equation}
where $H=\frac{X+Z-E-(Y_1+Y_2)/\mu}{2}$. It has closed-form solution according to singular value shrinkage, i.e., $D=Udiag((\sigma-\frac{1}{2\mu})_+)V^T$, $Udiag(\sigma)V^T$ is the singular value decomposition (SVD) of $H$.

Step 2: Updating $E$. We have
 \begin{equation}
\min_{E} \alpha\|E\|_1+\frac{\mu}{2}(\|E-(X-D-\frac{Y_1}{\mu})\|_F^2. \\
\label{solveE}
\end{equation}
It also admits close-formed solution, i.e., $e_{ij}=(|g_{ij}|-\alpha/\mu)_+\cdot sign(g_{ij})$, where $G=X-D-\frac{Y_1}{\mu}$.

Step 3: Updating $S$. Remember that $L$ is also a function of $S$, so the problem Eq. (\ref{lrgc}) becomes: 
\begin{equation}
\begin{split}
&\min_{s_i} \sum_{j=1}^n (\frac{\beta}{2}\| z_i-z_j\|^2s_{ij}+\gamma s_{ij}^2), \\
&\quad\hspace{.2cm} s.t. \hspace{.1cm} s_i^T\vec{1}=1, \hspace{.1cm}  0\leq s_{ij}\leq 1.
\end{split}
\label{sloves}
\end{equation}
Denote $f_{ij}=\|z_i-z_j\|^2$ and $f_i\in\mathcal{R}^{n\times 1}$, then the above subproblem can be reformulated as:
\begin{equation}
\min_{ s_i^T\vec{1}=1, 0\leq s_{ij}\leq 1} \|s_i+\frac{\beta}{4\gamma}f_i\|^2
\end{equation} 
The problem naturally has a sparse solution. Therefore, we just update its $k$ nearest neighbors, i.e., $s_i$ has $k$ positive entries and $s_{ij}=0$ for $j>k$. The Lagrangian function of it is:
\begin{equation}
\mathcal{L}(s_i,\eta,\xi)=\|s_i+\frac{\beta}{4\gamma_i}f_i\|^2-\eta( s_i^T\vec{1}-1)-\xi_i^Ts_i,
\end{equation}
where $\eta$ and $\xi\in\mathcal{R}^{n\times 1}$ are the Lagrangian multipliers and the overall $\gamma$ can be set to the average of $\{\gamma_i\}_{i=1}^n$. By the Karush-Kuhn-Tucker condition, it yields 
$s_i=(\frac{\eta}{2}-\frac{\beta f_i}{4\gamma_i})_+$. 

Let's rank $f_i$ in ascending order, we obtain
\begin{equation}
    \begin{cases}
s_{ik}=\frac{\eta}{2}-\frac{\beta f_{ik}}{4\gamma_i}>0
\vspace{.2cm}\\
 s_{i,k+1}\!=\!\frac{\eta}{2}\!-\!\frac{\beta f_{i,k+1}}{4\gamma_i}\!\leq\! 0\!
\vspace{.2cm}\\
s_i^T\vec{1}\!=\!\sum\limits_{j=1}^k\! (\!\frac{\eta}{2}\!-\!\frac{\beta f_{ij}}{4\gamma_i})\!=\!1\!
    \end{cases}
\label{updates}
\hspace{-.4cm}
\Rightarrow
    \begin{cases}
s_{ij}\!=\!\frac{f_{i,k+1}-f_{ij}}{kf_{i,k+1}\!-\!\sum\limits_{r=1}^k\! f_{ir}},\hspace{.05cm} j\leq k 
\vspace{.2cm}\\
\gamma_i=\frac{\beta}{4}(kf_{i,k+1}- \sum\limits_{j=1}^k f_{ij})
\vspace{.2cm}\\
\eta=\frac{2}{k}+\frac{\beta}{2k\gamma_i} \sum\limits_{j=1}^k f_{ij}
    \end{cases}
  \end{equation}
In the above derivations, we set the value of $\gamma_i$ to its maximum. Then, we take the average of $\{\gamma_i\}_{i=1}^n$, we have
\begin{equation}
\gamma=\frac{\beta}{4n}\sum\limits_{i=1}^n (kf_{i,k+1}- \sum\limits_{j=1}^k f_{ij}).
\label{gamma}
\end{equation}
 This ensures that the average number of nonzero elements in each row of $S$ is close to $k$.

Step 4: Updating $Z$. We have  
\begin{equation}
\min_Z \beta Tr(ZLZ^T)+\frac{\mu}{2}\|D-Z+\frac{Y_2}{\mu}\|_F^2). 
\end{equation}
Its first-order derivative is $2\beta ZL-\mu(D-Z+\frac{Y_2}{\mu})$.
By setting it to zero, we achieve
\begin{equation}
Z=(\mu D+Y_2)(2\beta L+\mu I)^{-1}.
\label{updateZ}
\end{equation}
The details of the algorithm are given in Algorithm 1. As we can see, the main computation load of this algorithm is from SVD involved in Eq. (\ref{solveD}) and matrix inversion in Eq. (\ref{updateZ}). Their computational complexity is $O(n^3)$ in general. Some existing packages can be employed to speed up their computation, e.g., partial SVD.
\begin{algorithm}
 \small  
\caption{The Proposed Method RGC}
\label{alg1}
 {\bfseries Input:} Data matrix $X$, parameters $\alpha>0$, $\beta>0$, $\gamma$, $\mu>0$.\\
{\bfseries Initialize:} $Z=X$, $E=0$, $Y_1=Y_2=0$.\\
 {\textbf{While} not converge \textbf{do}}
\begin{algorithmic}[1]
\STATE Calculate $D$ by (\ref{solveD}).
 \STATE Update $E$ according to (\ref{solveE}).
\STATE Update $S$ using (\ref{updates}) 
\STATE Calculate $Z$ using (\ref{updateZ}).
\STATE Update Lagrange multipliers $Y_1$ and $Y_2$ as
\begin{eqnarray*}
Y_1=Y_1+\mu(D+E-X),\\
Y_2=Y_2+\mu(D-Z).
\end{eqnarray*}
\end{algorithmic}
\textbf{End while.}
\end{algorithm}

\captionsetup{position=top}
\begin{table}[!htbp]
\centering
\caption{Description of the data sets}
\label{data}
\renewcommand{\arraystretch}{1.1}
\begin{tabular}{|l|c|c|c|}
\hline
&\textrm{\# instances}&\textrm{\# features}&\textrm{\# classes}\\\hline
\textrm{YALE}&165&1024&15\\\hline
\textrm{JAFFE}&213&676&10\\\hline
\textrm{ORL}&400&1024&40\\\hline
\textrm{COIL20}&1440&1024&20\\\hline
\textrm{BA}&1404&320&36\\\hline
\textrm{TR11}&414&6429&9\\\hline
\textrm{TR41}&878&7454&10\\\hline
\textrm{TR45}&690&8261&10\\\hline
\end{tabular}
\end{table}
\section{Experiments on Clustering}
\label{clustering}
In this section, we demonstrate the effectiveness of our robust graph construction technique towards clustering task. We compare it with several representative methods on a number of benchmark data sets.
\subsection{Comparison Methods} 
\begin{itemize}
\item \textbf{Spectral Clustering (SC) }\cite{ng2002spectral}. SC is a widely used clustering technique. It can separate different shaped groups. However, how to build a good graph is still a challenge. As we show later, we manually construct 12 graphs. For our proposed RGC method, we obtain clustering results by performing spectral clustering with our learned $S$.
\item\textbf{Robust Kernel K-means (RKKM)} \cite{du2015robust}. As an extension to classical $k$-means clustering method, RKKM has the capability of dealing with nonlinear structures, noise, and outliers in the data. RKKM shows promising results on a number of real-world data sets. 
\item \textbf{Constructing Robust Affinity Graph for Spectral Clustering (RSC)} \cite{zhu2014constructing}. In contrast to employing all features to construct a graph, RSC builds the graph via identifying and exploiting discriminative features by using random forest approach.
\item\textbf{Simplex Sparse Representation (SSR)} \cite{huang2015new}. Based on the self-expressiveness property, SSR constructs a sparse graph for clustering. It achieves satisfying performance in numerous data sets. 
\item \textbf{Clustering with Adaptive Neighbor (CAN)} \cite{nie2014clustering}. Based on the idea of adaptive neighbor, CAN learns a graph from raw data for clustering task. Different from CAN, our method learns the graph from clean data.
\item\textbf{Twin Learning for Similarity and Clustering (TLSC)} \cite{kang2017twin}: Based on self-expressiveness, TLSC has been proposed recently and has shown superior performance on a number of real-world data sets. TLSC learns graph matrix in kernel space. Since kernels are built from the original data, this method is also sensitive to noise. 
\item \textbf{Manifold RPCA (MRPCA)} \cite{shahid2015robust}. Rather than learn the graph from clean data, MRPCA constructs the graph from raw data according to some metrics. $K$-means is implemented on $D$ to achieve cluster labels.
\item Our proposed \textbf{RGC}\footnote{https://github.com/sckangz/RGC}. We learn reliable graphs from raw data, which is often corrupted in reality. 
\end{itemize}
\begin{table*}[!ht]
\centering
\renewcommand{\arraystretch}{1.1}
\setlength{\tabcolsep}{.1pt}
\subfloat[Accuracy(\%)\label{acc}]{
\resizebox{1.0\textwidth}{!}{
\begin{tabular}{ |l  |c |c| c| c |c| c| c| c| c| c| c|c  }
\hline
Data 	& RKKM\cite{du2015robust}		   & SC\cite{ng2002spectral}		      & RSC\cite{zhu2014constructing}	 &SSR\cite{huang2015new}     &CAN\cite{nie2014clustering}			         &TLSC\cite{kang2017twin} 	&MRPCA\cite{shahid2015robust}	      &RGC	   \\	\hline
YALE  	& 48.09 & 49.42 &57.58&54.55 & 58.79&55.85 &62.42      & \textbf{64.85} \\	\hline
JAFFE 	& 75.61 & 74.88 & 98.59&87.32 & 98.12 & \textbf{99.83}&98.12  & 98.59          \\	\hline
ORL   	& 54.96& 57.96 &62.75&69.00   & 61.50 &62.35  &56.75     & \textbf{73.00}	\\	\hline
COIL20 & 61.64 & 67.60 &72.99& 76.32  & 84.58 & 72.71  &73.47	  &\textbf{85.42}	\\	\hline
BA 		& 42.17& 31.07& 44.01 &23.97 &36.82 & 47.72    &44.16   	 & \textbf{51.00}	\\	\hline
TR11   	& 53.03 & 50.98 &34.54 &41.06& 38.89&71.26  &33.82     &\textbf{ 71.37 }    	\\	\hline
TR41  	& 56.76 & 63.52 &60.93 &63.78  & 62.87& 67.43&34.97	     & \textbf{70.16}     	\\	\hline
TR45  	& 58.13 & 57.39 &48.41 &71.45 & 56.96 & 74.02   &31.88   	 & \textbf{75.51}	\\	\hline
\end{tabular}
}
}\\
\renewcommand{\arraystretch}{1.1}
\subfloat[NMI(\%)\label{NMI}]{
\resizebox{1.0\textwidth}{!}{
\begin{tabular}{ |l  |c |c| c| c |c| c| c| c| c| c| c|c  }
\hline
Data 	& RKKM\cite{du2015robust}		   & SC\cite{ng2002spectral}		      & RSC\cite{zhu2014constructing}	 &SSR\cite{huang2015new}     &CAN\cite{nie2014clustering}			         &TLSC\cite{kang2017twin} 	&MRPCA\cite{shahid2015robust}	      &RGC	   \\	\hline
YALE 	& 52.29  & 52.92  &60.25&57.26 & 57.67 & 56.50    &62.86       & \textbf{65.29}	\\	\hline
JAFFE 	& 83.47 & 82.08  & 98.16&92.93   &97.31 & \textbf{99.35}  &97.72          & 98.13	\\	\hline
ORL 	& 74.23  & 75.16  & 79.87&84.23 & 76.59 & 78.96	&76.11         & \textbf{84.35}	\\	\hline
COIL20 & 74.63 & 80.98 &83.91& 86.89  & 91.55 & 82.20  &83.73  	  &\textbf{91.58}	\\	\hline
BA  	& 57.82 & 50.76  & 58.17 &30.29 & 49.32 & 63.04       &57.57          & \textbf{64.89}	\\	\hline
TR11  	& 49.69 & 43.11 & 24.77&27.60 &19.17 & 58.60&8.10&\textbf{67.74}	\\	\hline
TR41 	& 60.77  & 61.33 & 56.78&59.56 & 51.13 &65.50  &26.50   	&\textbf{67.35}   	\\	\hline
TR45	& 57.86  & 48.03  & 43.70&67.82   & 49.31 &\textbf{74.24 }    &17.81       &71.74	\\	\hline
\end{tabular}
}
}\\
\renewcommand{\arraystretch}{1.1}
\subfloat[Purity(\%)\label{Purity}]{
\resizebox{1.0\textwidth}{!}{
\begin{tabular}{ |l  |c |c| c| c |c| c| c| c| c| c| c|c  }
\hline
Data 	& RKKM\cite{du2015robust}		   & SC\cite{ng2002spectral}		      & RSC\cite{zhu2014constructing}	 &SSR\cite{huang2015new}     &CAN\cite{nie2014clustering}			         &TLSC\cite{kang2017twin} 	&MRPCA\cite{shahid2015robust}	      &RGC	   \\	\hline
YALE 	& 49.79  & 51.61  &63.03&58.18 & 59.39 & 56.50    &57.27       & \textbf{66.48}	\\	\hline
JAFFE 	& 71.82 & 76.83  & 97.65&96.24   &98.12 & \textbf{99.85}  &98.12          & 98.13	\\	\hline
ORL 	& 59.60  & 61.45  & 79.87&76.50 & 68.50 & 74.00	&62.50         & \textbf{76.75}	\\	\hline
COIL20 & 66.35 & 69.92 &80.83& 89.03  & 87.85 &82.53   &80.28  	  &\textbf{90.90}	\\	\hline
BA  	& 45.28 & 45.28  & 41.95 &40.85 & 39.67 & 52.36      &50.29          & \textbf{57.69}	\\	\hline
TR11  	& 67.93 & 67.93 & 35.75&85.02 &44.20 & \textbf{82.85}&62.32&79.85	\\	\hline
TR41 	& 74.99  & 74.99 & 55.58&75.40 & 67.54 &73.23  &41.80   	&\textbf{74.03}   	\\	\hline
TR45	& 68.18  & 68.18  & 45.51&83.62   & 60.87 &\textbf{78.26 }    &55.65       &77.39	\\	\hline
\end{tabular}
}
}\\
\caption{Clustering results measured on benchmark data sets. The best results for these algorithms are highlighted in boldface. \label{clusterres}}
\end{table*}
\subsection{Data Sets}
To fully examine the performance of our algorithm, we implement clustering experiments on eight publicly available data sets. We summarize the statistics information in Table \ref{data}. Specifically, YALE, JAFFE, and ORL are face databases, COIL20 is a toy image database, and BA is a binary alpha digits data set. They are taken under different illumination conditions or with different configurations, so some of them are corrupted severely. The last three data sets are text corpora derived from TERC\footnote{http://www-users.cs.umn.edu/ han/data/tmdata.tar.gz}. 
\subsection{Evaluation Metrics}

Three popular metrics are used to quantitatively assess our algorithm's performance on clustering task. They are accuracy (Acc), normalized mutual information (NMI), and Purity. 

Acc measures the one-to-one relationship between clusters and classes. Let $l_i$ and $\hat{l}_i$ be the algorithm's output and the ground truth label of $x_i$, respectively. Then the Acc can be defined as
\[
Acc=\frac{\sum_{i=1}^n \delta(\hat{l}_i, map(l_i))}{n},
\]
where $n$ denotes the number of samples, $\delta(\cdot)$ is the delta function, and map($\cdot$) is the best permutation mapping function. Based on Kuhn-Munkres algorithm, map($\cdot$) maps each cluster index to a true class label based.

The NMI is designed to measure the quality of clustering. Let $L$ and $\hat{L}$ denote two sets of clusters, then their marginal probability distribution functions can be represented by $p(l)$ and $p(\hat{l})$, respectively, induced from the joint distribution $p(l,\hat{l})$. Finally, the NMI can be defined by
\[
\textrm{NMI}(L,\hat{L})=\frac{\sum\limits_{l\in L,\hat{l}\in\hat{L}} p(l,\hat{l})\textrm{log}(\frac{p(l,\hat{l})}{p(l)p(\hat{l})})}{\textrm{max}(H(L),H(\hat{L}))},
\]  
The greater NMI means the better clustering performance.

The last evaluation metric that we adopt is the purity, which evaluates the extent to which the most common category in each cluster. It is computed as follows:
\[
\textrm{Purity}=\sum_{i=1}^{c} \frac{n_i}{n}P(C_i),\quad P(S_i)=\frac{1}{n_i}max_j(n_i^j), 
\]
where $n_i$ is the number of points in cluster $C_i$, and $n_i^j$ represents the total number of points that the $i$-th input group is assigned to the $j$-th category. There are $c$ categories in total. It is easy to see that a larger Purity indicated better clustering performance.

\subsection{Setup}
\label{setup}
For RKKM, SC, TLSC, MRPCA methods, we prepare their inputs by following the setup in \cite{du2015robust}. In particular, we design 12 kernels. They are: seven Gaussian kernels of the form $K(x,y)=exp(-\|x-y\|_2^2/(td_{max}^2))$, where $d_{max}$ is the maximal distance between samples and $t$ varies over the set $\{0.01, 0.0, 0.1, 1, 10, 50, 100\}$; a linear kernel $K(x,y)=x^T y$; four polynomial kernels $K(x,y)=(a+x^T y)^b$ with $a\in\{0,1\}$ and $b\in\{2,4\}$. Besides, all kernels are rescaled to $[0,1]$ by dividing each element by the largest pairwise squared distance. Then we report their best performances in Table \ref{clusterres}.

There are three parameters in our model. According to RPCA, we set $\alpha=1/\sqrt{max(m,n)}$ for all our experiments. And $\gamma$ is computed by (\ref{gamma}) in every iteration. In all experiments, we set neighbor number $k=10$. As a result, the only unknown parameter is $\beta$, and for this we can use cross validation.

\subsection{Results}
 We evaluate the clustering performance in terms of three widely used metrics: accuracy (Acc), normalized mutual information (NMI), and Purity. As can be seen from Table \ref{clusterres}, our method outperforms other state-of-the-art techniques in most experiments. Specifically, we have the following observations: 1) Our proposed method can beat CAN in all cases. Note that RGC constructs graph from clean data, while CAN learns graph from raw data. In particular, RGC improves the Acc of CAN from 36.83\%, 38.89\%, 56.96\% to 51\%, 71.37\%, 75.51\% on BA, TR41, TR45, respectively. This fully demonstrates the importance of constructing a robust graph; 2) Compared with self-expressiveness based graph construction methods SSR and TLSC, our method RGC achieves better performance in most cases. However, the performance of CAN is not as good as SSR and TLSC in most experiments. This also verifies the importance of removing noise in the data; 3) RGC performs much better than MRPCA in all cases. This is attributed to our adaptive graph learning strategy; 4) Compared with RKKM and SC, we also observe considerable improvements, although they have tried different kinds of kernels and RKKM is supposed to cope with noise; 5) With respect to RSC, RGC also has big advantages in terms of  Acc, NMI and Purity. To summarize, our proposed graph construction method can improve clustering performance significantly w.r.t. several newly developed techniques, this is mainly due to the removal of noise and errors.

\begin{figure}
\includegraphics[width=0.9\textwidth]{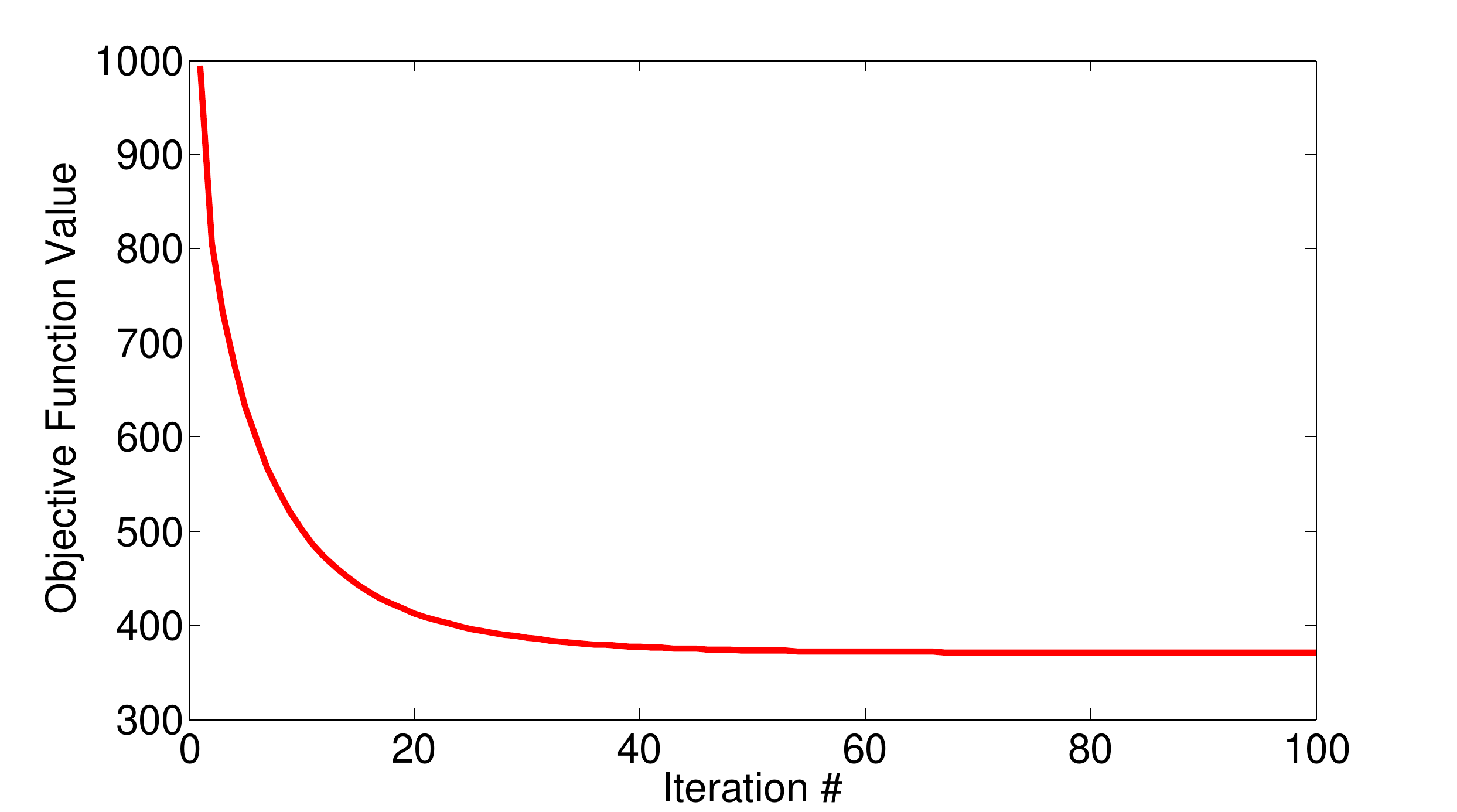}
\caption{Convergence curve of the objective function value (\ref{rgc}). }
\label{convergence}
\end{figure}

\begin{figure*}[!htbp]
\includegraphics[width=0.3\textwidth]{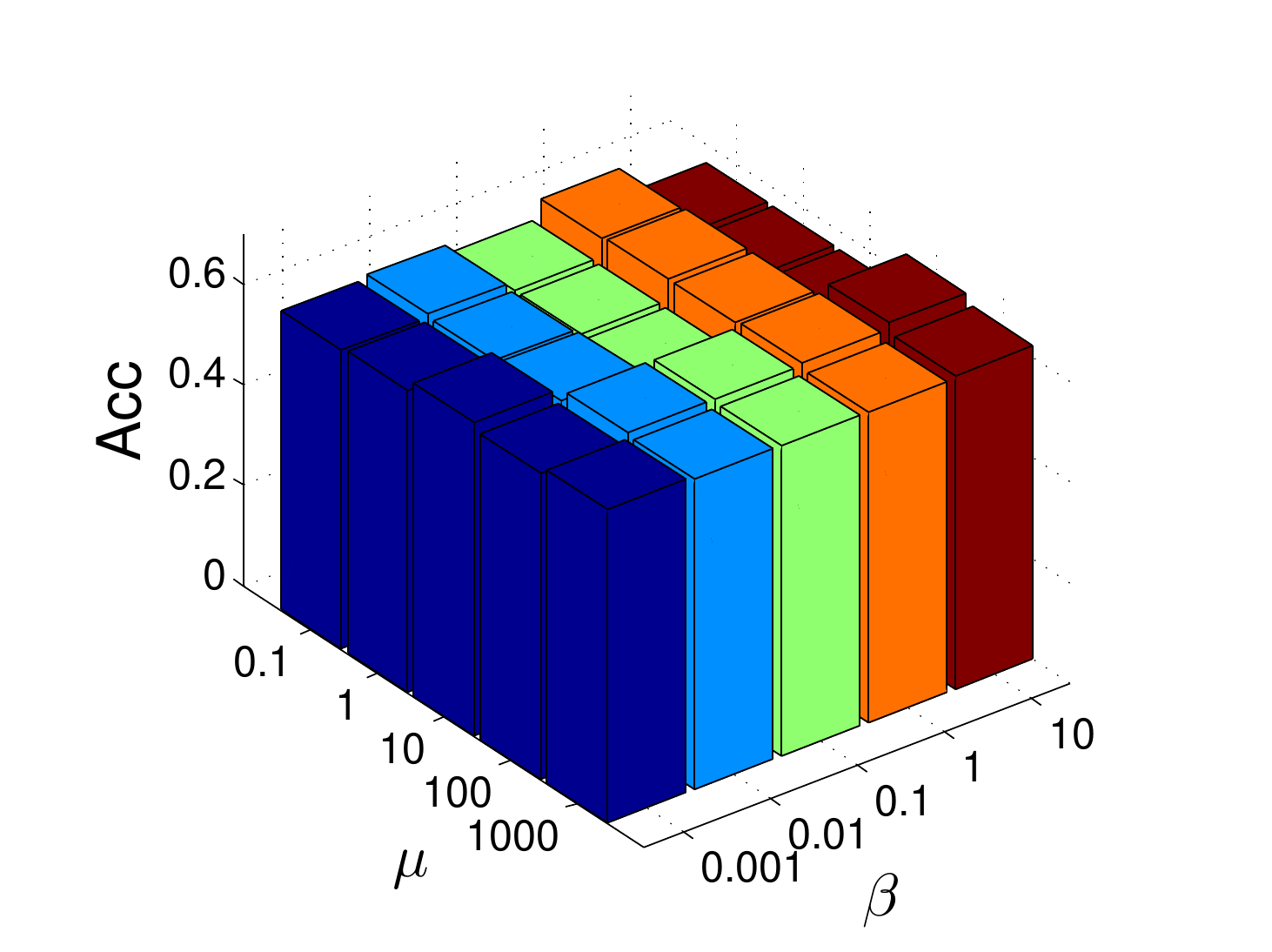}
\includegraphics[width=0.3\textwidth]{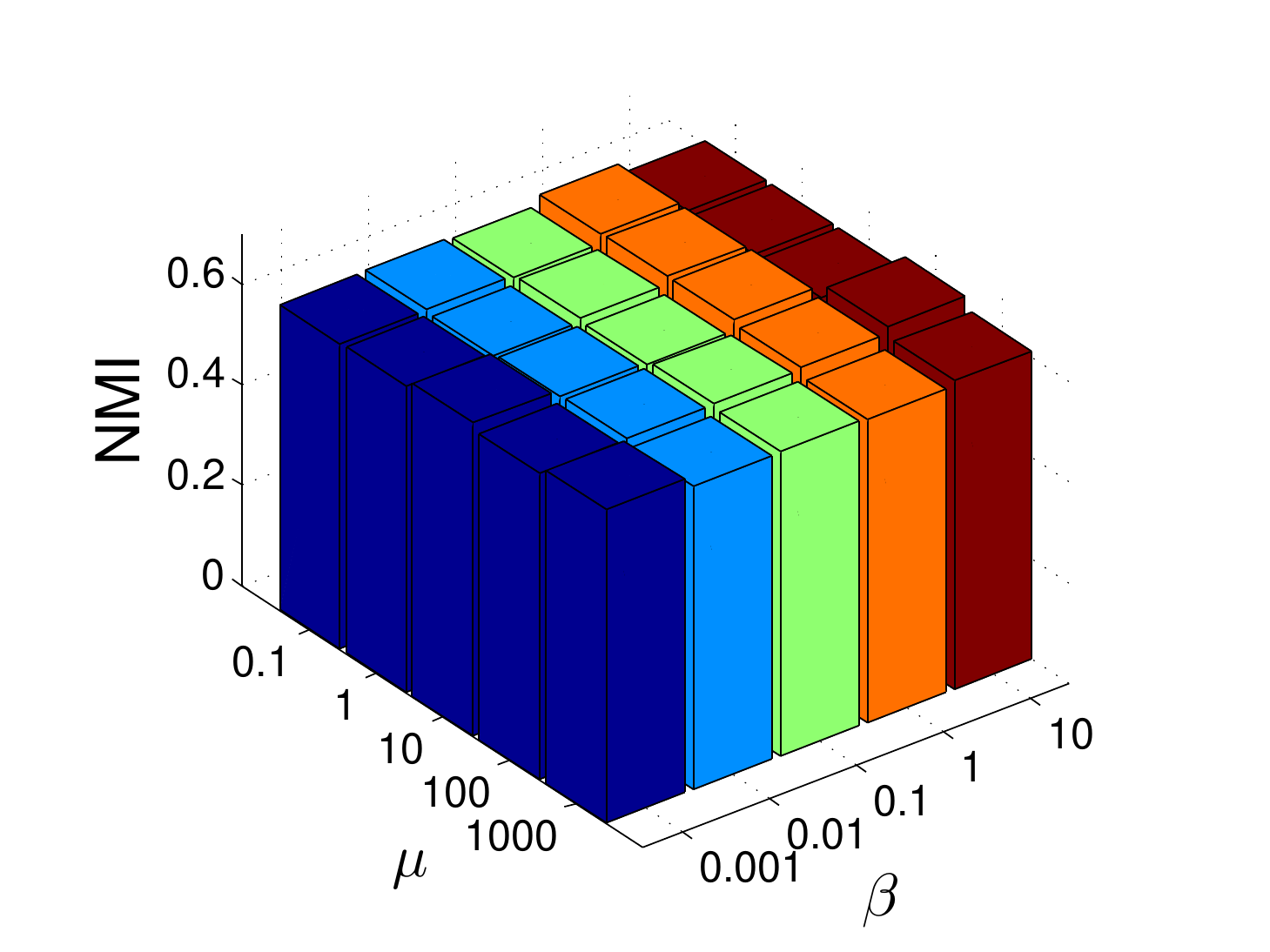}
\includegraphics[width=0.3\textwidth]{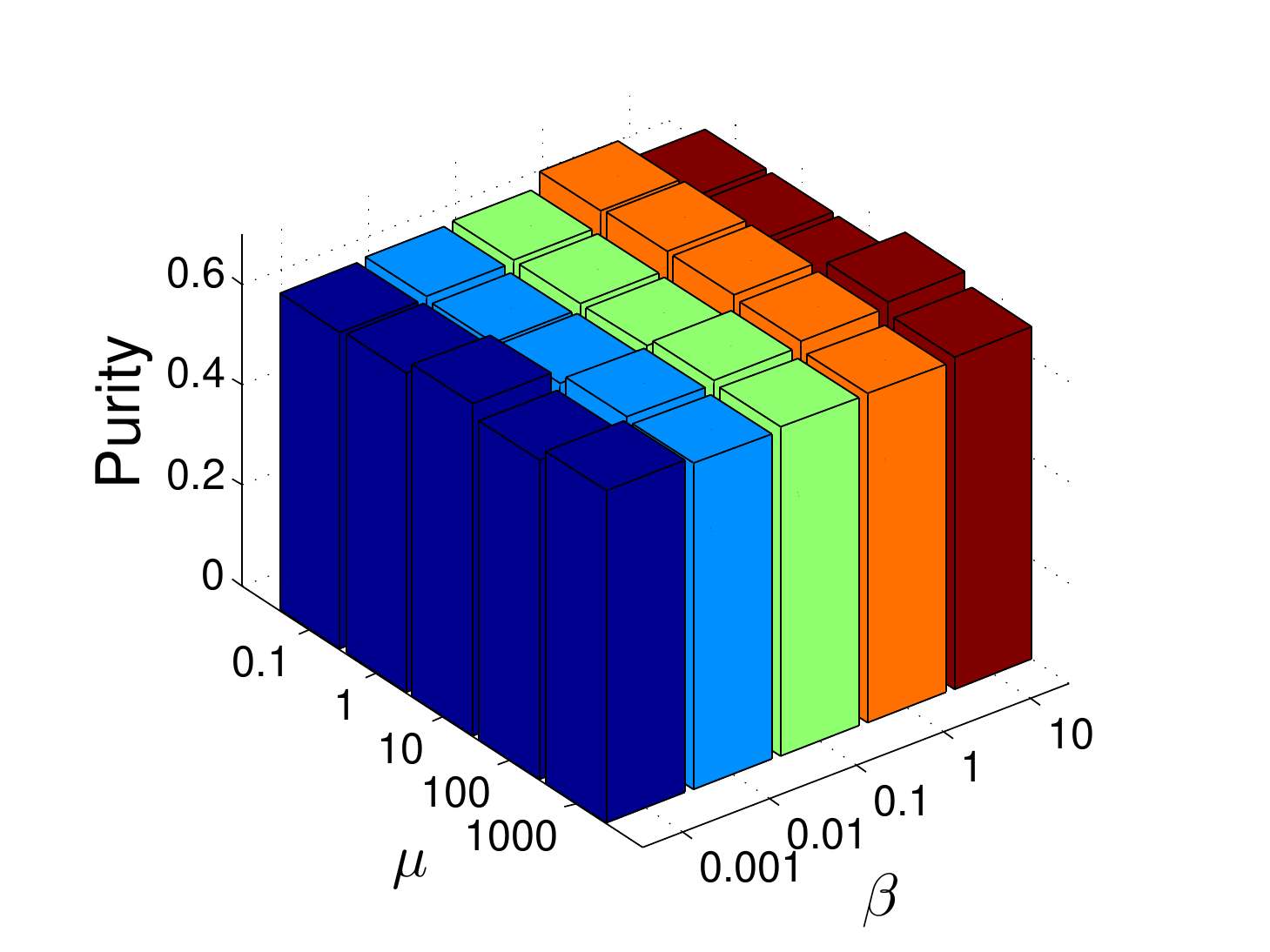}
\caption{The influence of $\beta$ and $\mu$ values on accuracy, NMI, and Purity of YALE data set. }
\label{parameter}
\end{figure*}

\subsection{Parameter Analysis}
As discussed in subsection \ref{setup}, model parameter $\beta$ and auxiliary parameter $\mu$ are unknown in our algorithm. Take YALE data set as an example, we show the sensitivity of our performance to their values. We can observe that Acc, NMI, Purity behave similarly with varying $\beta$ and $\mu$. Our method works well for a wide range of $\beta$ and $\mu$ values. Moreover, we show the progress of objective function values of Eq. (\ref{rgc}) in Figure \ref{convergence}. It is observed that our algorithm converges very fast.

\section{Semisupervised Classification }
\label{class}
\captionsetup{position=bottom}
\begin{figure*}[!htbp]
\centering
\subfloat[YALE\label{yale}]{\includegraphics[width=.28\textwidth]{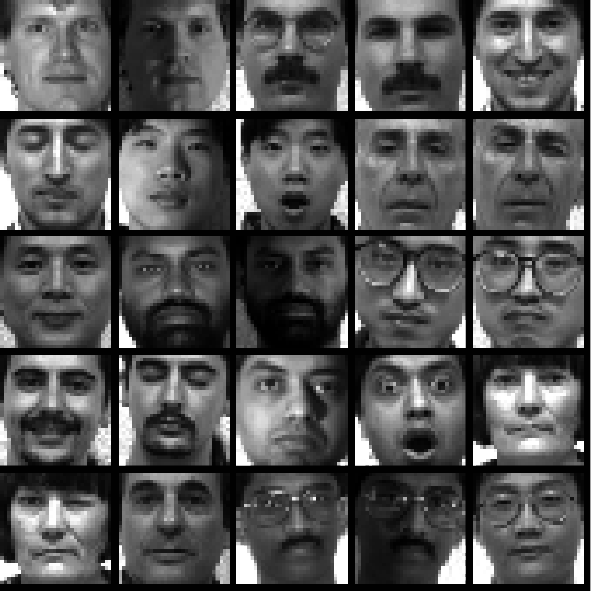}}
\subfloat[JAFFE \label{ba}]{\includegraphics[width=.28\textwidth]{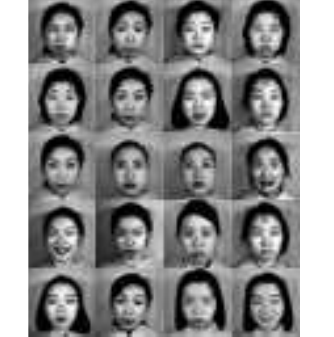}}
\subfloat[COIL20\label{coil20}]{\includegraphics[width=.36\textwidth]{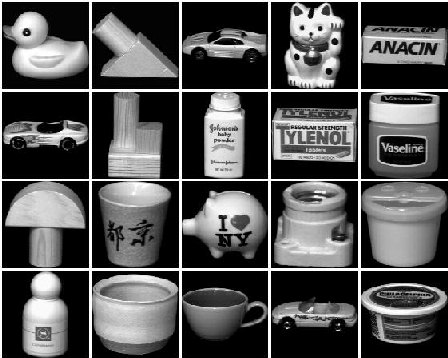}}
\caption{Sample images of YALE, JAFFE, and  COIL20.}
\end{figure*}

In this section, we show that our method also performs well on semisupervised classification task.
We first obtain the graph matrix $S$ from RGC, then we use the popular local and global consistency (LGC) method to perform classification task \cite{zhou2004learning}. Specifically, LGC achieves a classification function $F\in \mathcal{R}^{n\times c}$ by solving the following problem:
\begin{equation}
\min_F Tr\{F^T LF+\gamma(F-Y)^T(F-Y)\},
\end{equation} 
where $c$ is the class number, $Y\in \mathcal{R}^{n\times c}$ is the label matrix, in which $y_{ij}=1$ iff the $i$-th sample belongs to the $j$-th class, and $y_{ij}=0$ otherwise. As before, $L$ is the graph Laplacian matrix calculated from $S$. 

\subsection{Data Sets}
We evaluate our method on two types of recognition tasks: face recognition and visual object recognition. In specific,
we examine the effectiveness of our robust graph learning for face recognition on frequently used YALE and JEFFE data sets. The YALE face database has 15 individuals, and each individual has 11 near frontal images which are taken under different lighting conditions. Figure \ref{yale} displays some sample photos. The JAFFE face data set contains 10 individuals, and each person shows 7 different facial expressions, including 6 basic facial expressions and 1 neutral. We use COIL20 data set to perform visual object recognition experiment. It contains 20 objects and there are 72 photos for each object. For this data set, the images are taken 5 degrees apart when the object is rotating on a turntable. Figure \ref{coil20} shows some images from this database.\\

\subsection{Comparison Algorithms}
We compare RGC with five existing methods.
\begin{itemize}
\item {\textbf{Local and Global Consistency (LGC)} \cite{zhou2004learning}: LGC is a widely used semisupervised classification method.  }
\item{\textbf{Gaussian Field and Harmonic function (GFHF)} \cite{zhu2003semi}: Besides LGC, GFHF is another popular label propagating technique.}
\item{\textbf{Semisupervised Classification with Adaptive Neighbors (SCAN)} \cite{nie2017multi}: This recently developed method uses adaptive neighbors approach to construct the similarity graph. Moreover, the graph construction and clustering are formulated into a unified framework to improve the performance.}
\item{\textbf{Self-expression graph based Semisupervised Learning} \cite{li2015learning}: Different from SCAN, Li et al. propose to learn a graph for semisupervised learning by using self-expression method. Similar to SCAN, the similarity matrix and class indicator matrix $F$ are updated alternatingly, so they can help improve each other. Based on the low-rank and sparse assumption of the graph matrix, there are two models: S$^2$LRR and S$^3$R.}
\end{itemize}
Among these comparison techniques, both LGC and GFHF take $L$ as input. To obtain better performance, we calculate $L$ based on 7 kernel matrices. Then we report the best results from them. In specific, four Gaussian kernels with $t\in \{0.1, 1, 10, 100\}$; a linear kernel $K(x,y)=x^ y$; two polynomial kernels of the form with $K(x,y)=(a+x^T y)^2$ $a\in\{0,1\}$. For other techniques, the graph is learned from data.

\begin{table*}[htbp]
\begin{center}
\setlength{\tabcolsep}{1pt}
\renewcommand{\arraystretch}{1.5}
\caption{Classification accuracy (\%) on benchmark data sets (mean$\pm$standard deviation). The best results are in bold font.\label{classres}}
\resizebox{.99\textwidth}{!}{
\begin{tabular}{ |C{1.15cm}|C{2.1cm}|c|c|c|c|c|c|c|c|c|c| }
\hline
Data &Labeled Percentage($\%$) &GFHF\cite{zhu2003semi} & LGC\cite{zhou2004learning} &S$^3$R\cite{li2015learning}&S$^2$LRR\cite{li2015learning}& SCAN \cite{nie2017multi} & RGC\\
\hline\hline
\multirow{3}{4em}{YALE} & 10 &38.00$\pm$11.91&47.33$\pm$13.96& 38.83$\pm$8.60 &28.77$\pm$9.59& 45.07$\pm$1.30 & \textbf{58.90}$\pm$13.44\\ 
& 30 & 54.13$\pm$9.47&63.08$\pm$2.20& 58.25$\pm$4.25& 42.58$\pm$5.93& 60.92$\pm$4.03&\textbf{69.13}$\pm$3.84\\ 
& 50 & 60.28$\pm$5.16&69.56$\pm$5.42& 69.00$\pm$6.57& 51.22$\pm$6.78 & 68.94$\pm$4.57& \textbf{69.83}$\pm$5.95\\ 
\hline
\multirow{3}{4em}{JAFFE} & 10 & 92.85$\pm$7.76&96.68$\pm$2.76& 97.33$\pm$1.51& 94.38$\pm$6.23& 96.92$\pm$1.68 &\textbf{98.81}$\pm$0.72\\ 
& 30 &98.50$\pm$1.01&98.86$\pm$1.14& 99.25$\pm$0.81& 98.82$\pm$1.05& 98.20$\pm$1.22&\textbf{99.31}$\pm$0.81\\ 
& 50 &98.94$\pm$1.11&99.29$\pm$0.94& 99.82$\pm$0.60& 99.47$\pm$0.59 & 99.25$\pm$5.79&\textbf{99.87}$\pm$0.59\\ 
\hline\hline
\multirow{3}{4em}{COIL20} & 10 &87.74$\pm$2.26&85.43$\pm$1.40& 93.57$\pm$1.59& 81.10$\pm$1.69&90.09$\pm$1.15 &\textbf{95.71}$\pm$1.04\\ 
& 30 &95.48$\pm$1.40&87.82$\pm$1.03&96.52$\pm$0.68& 87.69$\pm$1.39 &95.27$\pm$0.93&\textbf{97.33}$\pm$0.92\\ 
& 50 &98.62$\pm$0.71&88.47$\pm$0.45&97.87$\pm$0.10& 90.92$\pm$1.19 &97.53$\pm$0.82&\textbf{99.80}$\pm$0.18\\ 
\hline
\end{tabular}}
\end{center}

\end{table*}

\begin{figure}[!ht]
\centering
\subfloat[Labeled Percentage ($10\%$)]{\includegraphics[width=.33\textwidth]{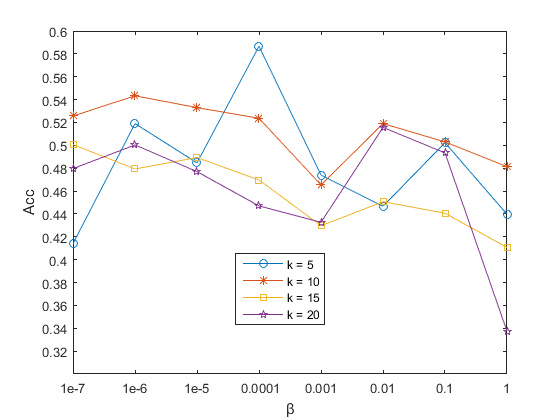}}
\subfloat[Labeled Percentage ($30\%$)]{\includegraphics[width=.33\textwidth]{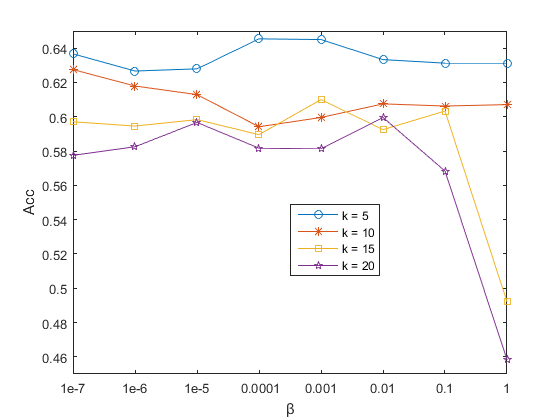}}
\subfloat[Labeled Percentage ($50\%$)]{\includegraphics[width=.33\textwidth]{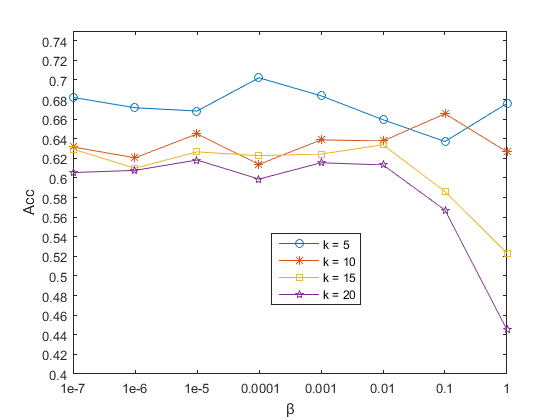}}
\caption{The influence of $\beta$ and neighbor number $k$ on YALE dataset.\label{parasense}}
\end{figure}

\subsection{Results}
To see the effect of label number, we randomly choose 10$\%$, 30$\%$, 50$\%$ of samples as labeled. We repeat this process 20 times and report the average classification accuracy and deviation in Table \ref{classres}. Unsurprisingly, the accuracy of all techniques goes up when the number of labeled samples increases. This is consistent with our intuition. As can be seen, our proposed RGC consistently outperforms other existing techniques. This demonstrates the importance of noise removal. Concretely, RGC improves on LGC significantly verifies the importance of graph construction. Please notice that LGC directly adopts manually built graph, while our method uses the learned graph $S$. Therefore, the enhancements are resulted from our high-quality graph construction. Another interesting observation about RGC is that the improvements are more considerable when the labeling ratio is relatively low. This makes RGC more attracting in real applications where labeled instances are extremely rare.

Though graph construction and label learning are implemented in a unified framework in SCAN, S$^2$LRR, and  S$^3$R, their performance is still not as good as our two-step approach. Once again, this is attribute to our graph learning strategy. Since SCAN and RGC are both based on adaptive neighbors graph construction, the superiority of RGC over SCAN is due to that RGC learns graph from clean data.
%
%
\subsection{Parameter Sensitivity}
We illustrate the parameter sensitivity for semisupervised learning task on YALE dataset in Figure \ref{parasense}. It is important to note that $k=5$ often produces the highest accuracy. Therefore, the results reported in Table \ref{clusterres} and \ref{classres}, where $k=10$ is adopted, could be further improved. Besides, our algorithm provides reasonable results for a wide range of $\beta$ value.

\begin{figure*}[!ht]
\centering
\begin{tabular}{cccccccc}
\hspace*{-7pt}\includegraphics[width=0.18\textwidth]{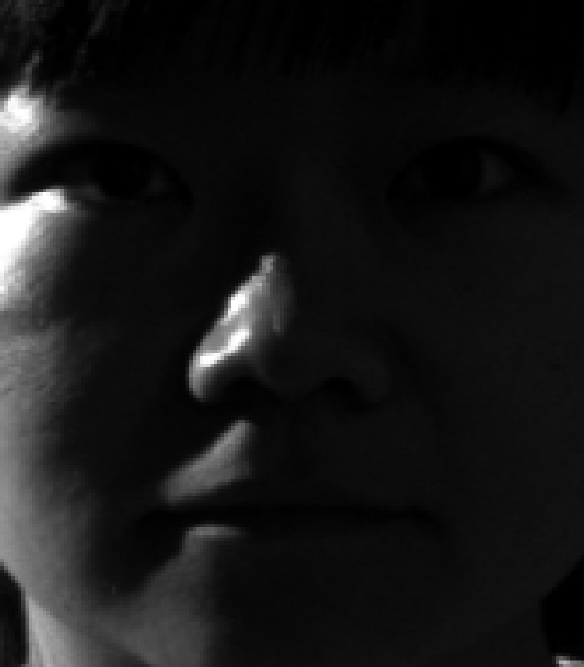}&
\hspace*{-5pt}\includegraphics[width=0.18\textwidth]{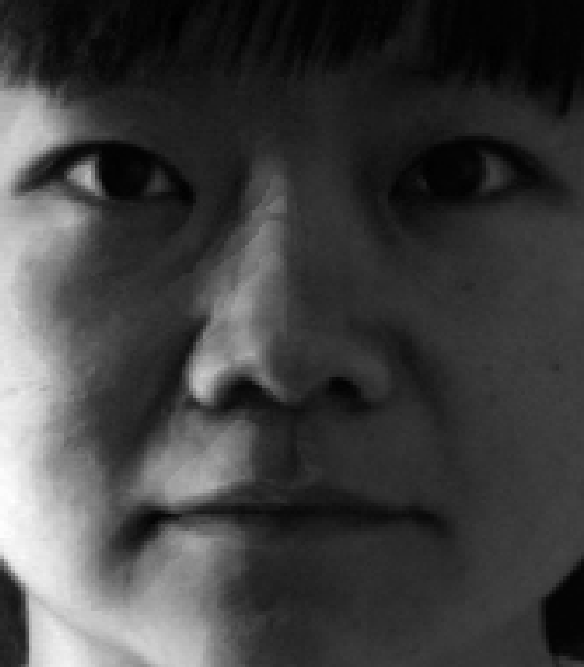}&
&
\hspace*{-5pt}\includegraphics[width=0.18\textwidth]{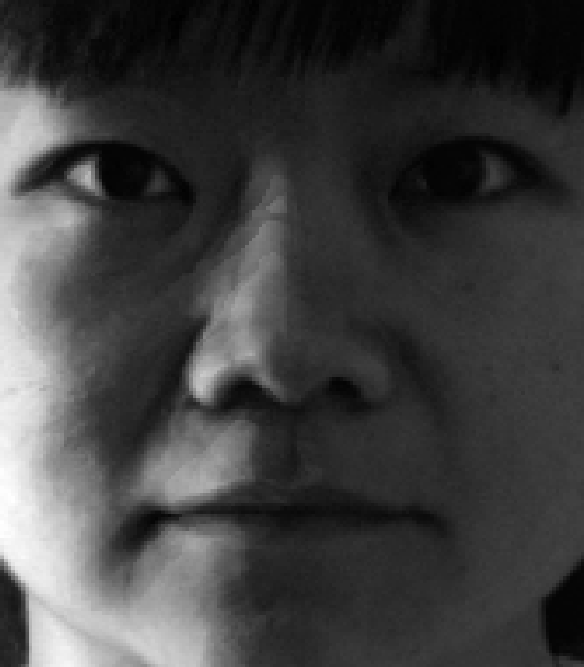}&
&
\hspace*{-5pt}\includegraphics[width=0.18\textwidth]{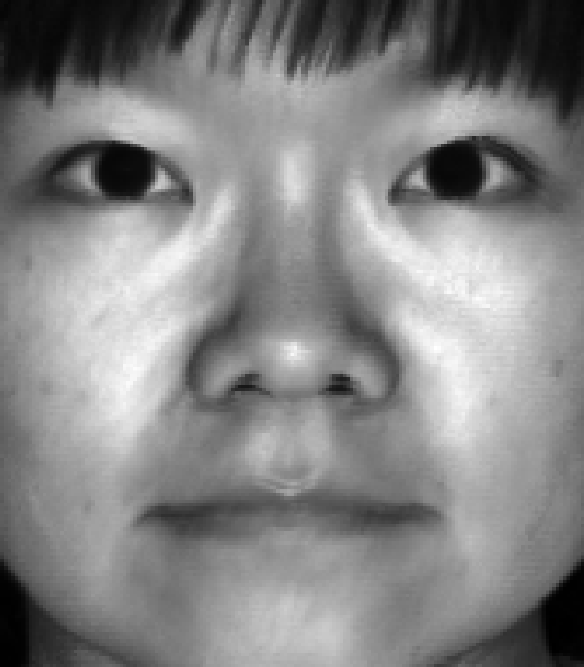}&
&
\hspace*{-5pt}\includegraphics[width=0.18\textwidth]{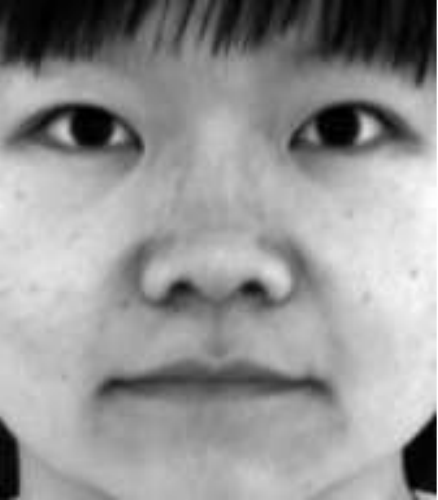}\\

 (a) Original images& (b) RPCA&&(c) CRPCA&&(d) NRPCA&&(e) RGC
\end{tabular}\vspace*{-1pt}
\caption{Shadow removal from face images. Column 1 displays sample images. Columns 2 to 5 show the recovered results obtained by different algorithms. }
\label{face}
\end{figure*}

\section{Experiments on Data Recovery}
\label{recovery}
In this section, we assess the performance of our model (\ref{rgc}) in low-rank matrix recovery task. RPCA has numerous applications. Due to space limitation, we focus on two practical tasks: shadow removal from face images and background extraction from videos.

\begin{figure*}[!h]
\centering
\begin{tabular}{cccccccc}
\hspace*{-7pt}\includegraphics[width=0.18\textwidth]{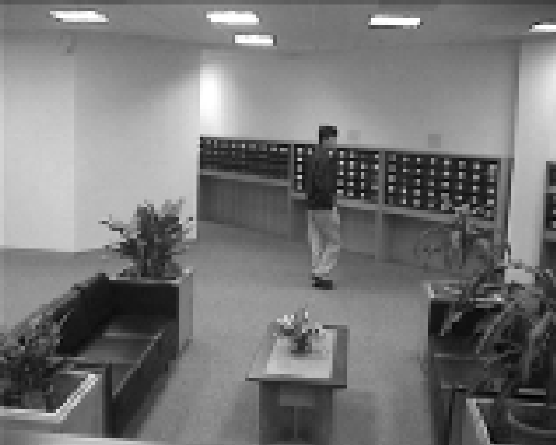}&
\hspace*{-5pt}\includegraphics[width=0.18\textwidth]{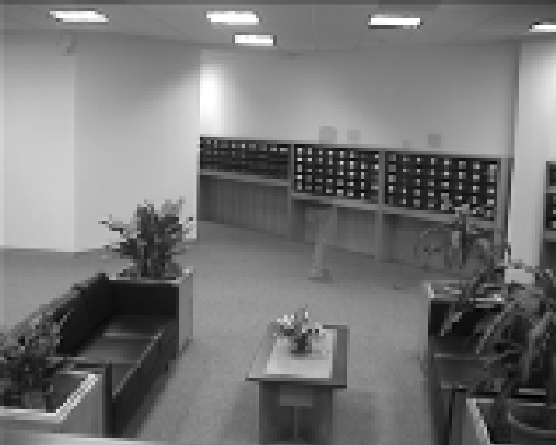}&
&
\hspace*{-5pt}\includegraphics[width=0.18\textwidth]{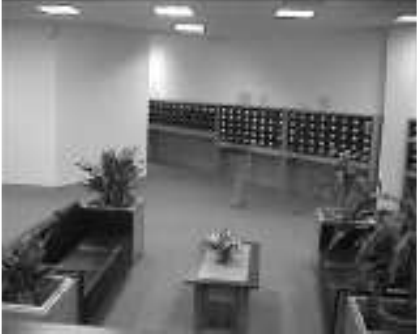}&
&
\hspace*{-5pt}\includegraphics[width=0.18\textwidth]{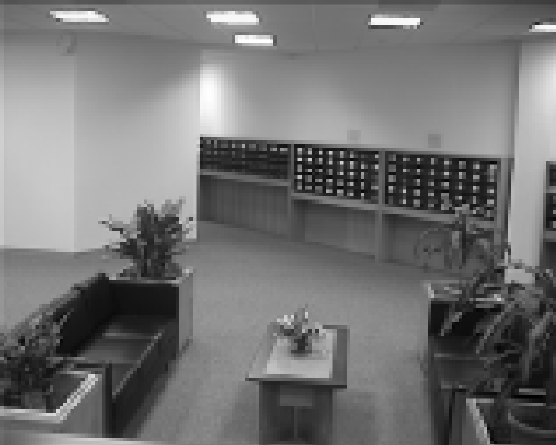}&
&
\hspace*{-5pt}\includegraphics[width=0.18\textwidth]{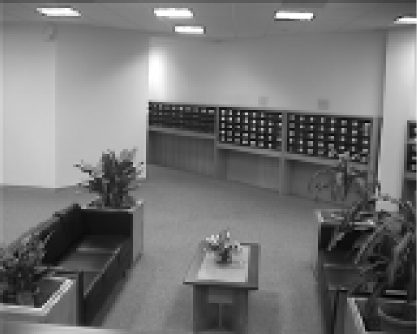}\\
 (a) Original images& (b) RPCA&&(c) CRPCA&&(d) NRPCA&&(e) RGC
\end{tabular}\vspace*{-1pt}
\caption{Low-rank background recovery. Column 1 displays actual frame. Columns 2 to 5 show the recovered low-rank background using different algorithms. }
\label{lobby}
\end{figure*}
\subsection{Comparison Methods} 
We choose a good set of methods to compare with our proposed method.
\begin{itemize}
\item \textbf{Robust Principle Component Analysis (RPCA)} \cite{candes2011robust}. Recovering clean data with low-rank structure from the corrupted data has been intensively studied during the last decade. One of the most representative techniques is RPCA. Low-rank part and sparse part can effectively characterize the clean data and errors, respectively.
\item\textbf{Capped-norm based RPCA (CRPCA)} \cite{sun2013robust}. RPCA uses the nuclear norm to approximate the rank function, which generally leads to an NP-hard problem. However, the nuclear norm shrinks all the singular values equally and thus over-penalizes large singular values, which often results in a biased solution. CRPCA adopts a truncation strategy to alleviate the nuclear norm deficiency and leads to better performance.
\item\textbf{Nonconvex RPCA (NRPCA)} \cite{netrapalli2014non}. By combining the simplicity of PCA and elegant theory of RPCA, NRPCA has been developed recently. It alternatingly projects the residuals onto the low-rank and sparse matrix sets. It has global convergence and exact recovery guarantee. However, it also requires the knowledge of sparsity, 
\item  Our proposed \textbf{RGC} method. Compared with RPCA, our method incorporates manifold smoothness effect. Both CRPCA and NRPCA adopt a nonconvex approach, while our method still uses convex nuclear norm.
\end{itemize}

\subsection{Face Image Shadow Removal}
Removing shadows, specularities, and saturations in face images is crucial to face recognition. With  a bunch of face images, we are supposed to recover the clean image. We randomly choose one subject from the Extended Yale B database. There are 64 images of size 192$\times$168 taken under different lighting conditions. All images are vectorized, hence $X\in \mathcal{R}^{32256\times 64}$.

Figure \ref{face} shows the recovered images from an example image. We can observe that our proposed RGC totally removes the specularities and shadows. And there are still some artifacts left for other three methods, especially for RPCA and CRPCA. 

\subsection{Background Extraction from Videos}

An important application of low-rank recovery is to separate static background from the dynamic foreground. We use benchmark data set lobby, which contains 1,546 frames of size $128\times 160$. After vectorizing and concatenating, the size of $X$ becomes $20,480\times 1,546$. Since the low-rank ground truth is not available, we present a visual comparison of recovered low-rank background in Figure \ref{lobby}. We can see that our method totally removes the walking person, while some shadows of pant are left in the frame recovered by RPCA, so the presence of graph in our model enables better recovery. For NRPCA, if we assume a small rank value (e.g., 2 here), we can also get a pretty clean frame.

\section{Conclusion}
\label{conclusion}
In this paper, we present a model to learn reliable graphs from raw data. It aims to address the robustness issue of adaptive neighbor graph learning method. From another point of view, it is a generalization of the Robust PCA by leveraging graph regularization on the low-rank representation. Consequently, our proposed framework can either enhance clustering/semisupervised classification or improve low-rank recovery from grossly corrupted data sets. Experiments on several benchmark data sets reveal that our method outperforms various state-of-the-art clustering/semisupervised classification and data recovery techniques.


\bibliographystyle{IEEEtran}
\bibliography{ref}

\end{document}